\documentclass[conference]{IEEEtran}

\usepackage{amsmath,amsfonts,amssymb}
\usepackage{graphicx,booktabs}
\usepackage{multirow,balance,url}

\title{Patch-Level Tokenization with CNN Encoders and Attention for Improved Transformer Time-Series Forecasting}

\author{
\IEEEauthorblockN{Saurish Nagrath, Saroj Kumar Panigrahy, and Soubhagya Sankar Barpanda}
\IEEEauthorblockA{\textit{School of Computer Science and Engineering, VIT-AP University, Amaravati, Andhra Pradesh, 522241, India} \\
* Correspondence: saroj.panigrahy@vitap.ac.in}   
}

\begin{document}
\maketitle
\begin{abstract}
Transformer-based models can capture long-range temporal dependencies, but forecasting performance depends on the quality of their input representations. This paper presents a two-stage framework that separates local temporal representation learning from global temporal modelling. A shared convolutional neural network (CNN) encodes non-overlapping temporal patches, dense connections preserve intermediate features, attention-weighted pooling aggregates within-patch information, and token-level self-attention refines patch embeddings before a Transformer encoder models inter-patch dependencies. The framework is evaluated on a controlled synthetic multivariate time-series dataset and the UCI Electricity Load Diagrams dataset. Synthetic experiments examine forecast horizons of one, three, and five patches. At the one-patch horizon, the proposed model achieves an MSE of 0.076 and an MAE of 0.205, compared with 0.095 and 0.218 for PatchTST and 0.238 and 0.355 for a temporal convolutional network (TCN). The proposed model also reports the lowest MSE and MAE at the longer evaluated horizons. On the electricity dataset, it obtains an MSE of 0.090665, an MAE of 0.142513, and an RMSE of 0.301105, outperforming the reported TCN and PatchTST baselines. An ablation study evaluates dense connections, attention-weighted pooling, and token-level self-attention. The results support the proposed representation strategy in the evaluated settings, while additional experiments on diverse real-world benchmarks remain necessary.
\end{abstract}


\section{Introduction}\label{i.-introduction}

Accurately forecasting multivariate time-series data remains a challenging problem due to the complex, noisy, and often non-stationary nature of many real-world temporal processes. Such time-series data are shaped by a mix of short-lived fluctuations, longer-term trends, and interactions between multiple variables. Together, these factors introduce non-linear behaviour, uneven feature dependencies, and temporal patterns that change over time. As a result, designing forecasting models that perform reliably across different application domains remains a challenging task.
Traditional time-series forecasting methods, such as autoregressive and moving-average models \cite{patel2015,chaudhary2025}, are built on assumptions of linearity and stationarity. In many real-world settings, these assumptions do not hold up, which limits the effectiveness of such an approach. This has led to growing interest in deep learning techniques for modelling complex temporal behaviour. Recurrent neural networks (RNNs) \cite{lindemann2021} have been among the most widely adopted of these methods, as they are specifically designed to process sequential data, particularly Long Short-Term Memory (LSTM) \cite{lindemann2021} and Gated Recurrent Unit (GRU) \cite{shen2018} architectures, are among the most widely adopted models due to their ability to process sequential data and capture local temporal dependencies. However, recurrent models summarize historical information through compressed hidden states, which can limit their ability to retain long-range dependencies and may lead to unstable training behaviour when long input sequences are considered.

Transformer-based architectures \cite{wen2022} have recently emerged as a powerful alternative for time-series forecasting. By employing self-attention mechanisms \cite{vaswani2017}, Transformers enable direct interactions between all elements of a sequence, removing the reliance on recursive state propagation. This allows more effective modelling of long-range temporal dependencies and improved parallelization during training. Several studies have reported promising results using Transformer models for time-series forecasting \cite{stockformer2025,quantformer}. Nevertheless, Transformer performance is highly sensitive to the structure and quality of the input representations provided to the attention mechanism.

Unlike text or image data, multivariate time-series do not possess a natural tokenization scheme. When Transformers are applied directly to raw sequences, the model must simultaneously learn low-level feature extraction, noise suppression, and high-level temporal relationships. In many time-series settings, predictive information is concentrated in localized temporal patterns, such as short-term dynamics and volatility-like behaviour driven by recent observations \cite{jegadeesh1993,engle1982}. Attempting to capture both local temporal structure and long-range dependencies within a single attention-based model can therefore lead to inefficient use of model capacity and suboptimal representation learning \cite{bai2018,nie2023}.

Motivated by recent advances in representation learning, which suggest that explicitly separating local pattern extraction from global dependency modelling can improve learning efficiency and performance \cite{long2025}, this work adopts a patch-based representation learning strategy for multivariate time-series forecasting. The proposed framework operates on fixed-length temporal patches, treating them as the fundamental units of representation. In the first stage, a convolutional neural network (CNN) encodes non-overlapping temporal patches to extract short-range temporal dynamics and non-linear feature interactions. Token-level self-attention is subsequently applied during representation learning to refine these patch embeddings by enabling interactions across temporal segments. In the second stage, a Transformer encoder processes the resulting sequence of patch-level tokens to model inter-patch temporal dependencies and generate forecasts.

The proposed architecture is initially evaluated on a constructed synthetic dataset designed to reflect key properties and structural dependencies as is observed in real world data, including static and dynamic features as well as stochastic noise. Performance is compared against convolutional and patch-based Transformer baselines: Temporal Convolutional Networks (TCN) \cite{bai2018} and PatchTST \cite{nie2023}, and results demonstrate that patch-level temporal tokenization consistently achieves competitive forecasting accuracy on unseen data. Model is then evaluated on a real dataset to show architecture capability in real scenarios and not just tailored situations. 

The main contributions of this work are fourfold. First, it separates local temporal feature extraction from global inter-patch forecasting in a two-stage architecture. Second, it combines dense CNN feature extraction with attention-weighted pooling and token-level self-attention to form structured patch embeddings. Third, it evaluates the approach against TCN and PatchTST across multiple synthetic forecast horizons and on electricity-load data. Fourth, it reports an ablation study to examine the contribution of individual representation components.

The remainder of this paper is organized as follows. Section~II
describes the proposed patch-based forecasting framework, detailing the
CNN-based patch encoder, token-level self-attention mechanism, and the
specialized Transformer architecture used for global temporal modelling.
Section~III presents the experimental setup, including datasets used and how synthetic dataset was generated, baseline models, implementation details, and hyperparameter
settings. Section~IV reports the empirical results and provides a
comparative analysis of the proposed method against the baseline approaches.
Section~V concludes the paper and draws subjective conclusions for the proposed model.
The paper presents the proposed architecture, experimental setup, results, and conclusion.

\section{Methodology}\label{ii.-method}

\subsection{Proposed method}\label{ii.i.-proposed-method}

Let $\mathbf{X} = \{x_t\}_{t=1}^{T}$, where $x_t \in \mathbb{R}^{F}$,
denote a multivariate time series of length $T$. Rather than applying a
forecasting model directly to the raw input sequence, the proposed
approach first transforms $\mathbf{X}$ into a sequence of learned
temporal tokens. The time series is partitioned into non-overlapping
temporal patches of fixed length $P$, resulting in
$K = \lfloor T / P \rfloor$ segments. Each patch is processed
independently by a shared convolutional neural network, which captures
short-range temporal patterns and non-linear interactions among features
within the patch. The convolutional representations are subsequently
aggregated along the temporal dimension and projected into a
$D$-dimensional embedding space, producing a sequence of patch-level
tokens $\mathbf{Z} \in \mathbb{R}^{K \times D}$. To enable information
exchange across temporal segments, multi-head self-attention is applied
to the token sequence during representation learning, allowing
dependencies between patches to be explicitly modelled. The refined token sequence is subsequently used as input to a Transformer encoder, which models inter-patch temporal dependencies and produces patch-level forecasts. This
design separates local representation learning from global temporal
modelling, leading to more stable and effective forecasting.
\begin{figure}[t]
\centering
\includegraphics[width=0.8\linewidth]{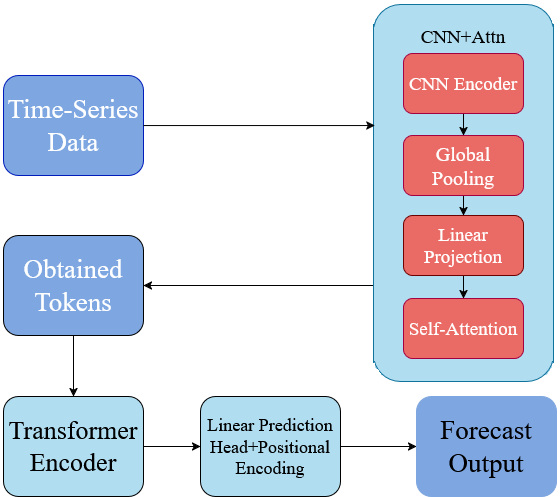}
\caption{Overview of the proposed CNN--Transformer architecture for multivariate time-series forecasting.}
\label{fig:architecture}
\end{figure}
\subsection{CNN Patch Encoder and Attention}\label{ii.ii.-self-attention-cnn}

The convolutional neural network is employed as a patch encoder used to
transform raw multivariate time-series input into optimized temporal
token embeddings. A convolutional neural network is employed in the first stage due to its strong inductive bias toward extracting localized and stable patterns, making it well suited for modelling static or slowly varying feature interactions in multivariate time-series data. Subsequent token-level self-attention is applied to the patch embeddings produced by the convolutional encoder, enabling interactions across temporal patches during token refinement, extending beyond the local convolutional receptive field \cite{enhancingtransformer}. Given an input sequence \(\mathbf X \in \mathbb R^{T\times F}\), where \(T\) is the sequence length and \(F\) is the number of input features, the sequence is first partitioned into non-overlapping fixed length temporal patches \(P\). This results in a sequence of \(K=\lfloor T/P \rfloor\) patches, each representing a localized temporal segment of the original series.

Each patch \(\mathbf X_k\in \mathbb R^{P\times F}\) is processed
independently using a shared one-dimensional convolutional encoder.
Convolutions are applied along the temporal dimension, treating the
feature dimension as channels, allowing the network to capture
short-term temporal patterns and non-linear interactions among features
within each patch. Dense connections are applied within the convolutional blocks of the encoder to preserve intermediate representations and promote future use, enabling the contribution of both low and higher-level temporal features in final representation.
Following the convolutional processing, a learned attention weighted pooling is applied across the temporal dimension of each patch to obtain a fixed length vector representation. This pooled representation is then projected into a \(D\)-dimensional embedding space, thus producing an initial token embedding for each patch. To incorporate contextual information across patches, multi-head self-attention is applied to the sequence of token embeddings. This allows each token to attend to others, refining the
representations by integrating information from neighbouring and distant
temporal segments. The output of this module is a sequence of optimized
temporal token embeddings, which serve as structured inputs for
subsequent Transformer-based modelling.

\subsection{Specialised
Transformer}\label{ii.iii.-optimized-transformer}

The Transformer stage receives as input a sequence of token embeddings
\(\mathbf{X} \in \mathbb{R}^{B \times K \times D_{\text{in}}}\), where
\(B\) is the batch size, \(K\) is the number of temporal tokens, and
\(D_{in}\) is the embedding dimension produced by the CNN encoder. If
the embedding dimension of the input differs from the model dimension
\(d_{\text{model}}\), a linear projection is first applied to map the
input tokens to a common latent space.

\[
\mathbf X'=\mathbf X\mathbf W_p + b_p
\]

\[
\mathbf W_p \in \mathbb R^{D_{in}\times d_{model}}
\]

To preserve temporal ordering, learned positional embeddings are added
to the projected tokens. For each temporal patch
\(k\in \{1,\dots,K\}\), a positional vector
\(\mathbf{p}_k \in \mathbb{R}^{d_{\text{model}}}\) is retrieved from an
embedding table and combined with the token representation,

\[
\mathbf {\tilde x}_k = \mathbf{x}'_k + \mathbf{p}_k
\]

Dropout is applied to the position-enhanced sequence to regularize
training.

The resulting sequence is processed by a stack of Transformer encoder
layers operating in a pre-normalization configuration. Each layer
applies multi-head self-attention followed by a position-wise
feed-forward network, with residual connections and layer normalization
to stabilize optimization. Self-attention enables each token to
aggregate information from all other tokens in the sequence, allowing
the model to capture long-range temporal dependencies.

Finally, a linear output head is applied to the Transformer outputs to
produce forecasts at a fixed patch-level horizon. Given a forecast
horizon \(h>0\), the model predicts future patch-level targets according
to

\[
\hat y_{k+h}=\mathbf{w}^{\top}\mathbf{h}_k+b,
\]

\noindent where \(\mathbf h_k\) denotes the Transformer output corresponding to
temporal patch \(k\). The Transformer is trained by minimizing a regression loss between the predicted sequence and the corresponding ground-truth patch-level targets, while the patch encoder remains fixed.

Ground-truth targets are aligned to the patch structure by aggregating the original time-series values within each patch using a fixed reduction operation.

\subsection{Prediction
Mechanism}\label{ii.iv.-prediction-mechanism}

After training, the model generates forecasts by mapping encoded
temporal representations to future patch-level target values. Given an
input sequence of length \(T\), the Transformer encoder produces a
sequence of hidden states
\(\{\mathbf{h}_k\}_{k=1}^{K}\), where each hidden state
\(\mathbf{h}_k \in \mathbb{R}^D\) corresponds to a specific temporal
patch and encodes information aggregated from all patches in the input sequence through self-attention. 

For a fixed forecast horizon \(h\), predictions are generated using
hidden states up to patch \(k\) to forecast the target at patch
\(k+h\). A linear output layer is applied to each hidden state to obtain
the predicted sequence,

\[
\hat{y}_{k+h} = \mathbf{w}^\top \mathbf{h}_k + b,
\quad k = 1,\dots,K-h,
\]

\noindent where \(\mathbf{w}\) and \(b\) are learned parameters. The model is
trained to minimize a regression loss between the predicted patch-level
forecasts \(\{\hat y_{k+h}\}\) and the corresponding ground-truth
patch-level targets. At inference time, the same horizon-aligned mapping is
applied to previously unseen input sequences, producing a sequence of
one-step-ahead patch-level forecasts aligned with the temporal patch
structure of the input.

\subsection{Patch-level target construction}
For a target sequence $y_1,\ldots,y_T$, the target associated with a complete patch is obtained by arithmetic mean reduction. For patch $j$ of length $P$, the target is
\begin{equation}
\bar{y}_j=\frac{1}{P}\sum_{r=1}^{P}y_{(j-1)P+r},\quad j=1,\ldots,\left\lfloor\frac{T}{P}\right\rfloor.
\label{eq:patch-target-mean}
\end{equation}
Only complete patches are retained; trailing time steps that do not fill a patch are discarded. This makes the forecasting target explicit and aligns predictions and targets at the same temporal granularity. The encoder and forecasting stage are optimized separately: the CNN encoder is pretrained for patch-level representation reconstruction, and the Transformer is subsequently trained for forecasting using the encoded patch sequence.

\section{Experimental setup}\label{iii.-experiment}

\subsection{Datasets}\label{iii.i.-dataset}
\subsubsection{Preparation}

The experiments are conducted on a synthetic multivariate time-series
dataset designed to provide controlled interactions between dynamic and
static features \cite{long2025}. Each sample consists of a time series of length $T$,
where each observation $x_t \in \mathbb{R}^6$ contains two dynamic
features and four static features.

The first dynamic feature $d_t^{(1)}$ is generated as a non-linear
function of time,
\[
d_t^{(1)} = f(t) + \epsilon_t^{(1)},
\]
where $f(t)$ combines polynomial, exponential, and sinusoidal components,
and $\epsilon_t^{(1)} \sim \mathcal{N}(0,\sigma_1^2)$ denotes Gaussian
noise.

The second dynamic feature $d_t^{(2)}$ is generated as a non-linear
function of the first dynamic feature together with additional
time-dependent components,
\[
d_t^{(2)} = \rho\, g(d_t^{(1)}) + h(t) + p(t) + \epsilon_t^{(2)},
\]
where \(g(\cdot)\) denotes a fixed non-linear transformation composed of
linear and logarithmic terms, $h(t)$ denotes a trend component, $p(t)$ denotes a
periodic component, $\rho$ controls the dependency strength, and
$\epsilon_t^{(2)} \sim \mathcal{N}(0,\sigma_2^2)$.

Four static features $\{s_1, s_2, s_3, s_4\}$ are sampled once per
sequence and remain constant across all time steps. These static
features are repeated along the temporal dimension and concatenated with
the dynamic features to form the input at each time step,
\[
x_t = [ d_t^{(1)}, d_t^{(2)}, s_1, s_2, s_3, s_4 ].
\]

The target sequence $\mathbf{Y} = \{ y_t \}_{t=1}^T$ is generated as a
non-linear combination of dynamic and static features,
\[
y_t =
\alpha_1 d_t^{(1)} s_1
+ \alpha_2 \frac{35 - s_2}{s_2 - 3}
+ \alpha_3 d_t^{(2)} s_4
+ \epsilon_t^{(y)},
\]
where $\alpha_1 = 0.045$, $\alpha_2 = 0.38$, $\alpha_3 = 0.07$, and
$\epsilon_t^{(y)} \sim \mathcal{N}(0,\sigma_y^2)$. The noise variances \(\sigma_1^2\), \(\sigma_2^2\), and \(\sigma_y^2\) are
fixed constants controlling the noise level in the synthetic data
generation process.

The resulting dataset consists of input tensors
$\mathbf{X} \in \mathbb{R}^{N \times T \times 6}$ and target tensors
$\mathbf{Y} \in \mathbb{R}^{N \times T}$. \\

\subsubsection{Seeding}
To ensure a fair and reproducible evaluation, two independent datasets were generated using random seeds 42 and 101 for training and testing, respectively. Both datasets follow identical data generation and preprocessing pipelines. All proposed and baseline models are trained exclusively on the training dataset and evaluated on the held-out test dataset.

\subsubsection{Illustrative Sample}

Table~\ref{tab:datasample} presents an illustrative example from the
synthetic dataset, showing the first few time steps of a single
multivariate time-series sample. Each observation consists of two
dynamic features and four static features, where the static features
remain constant across time steps.
\begin{table}[h]
\centering
\caption{Illustrative example of a single synthetic multivariate
time-series sample (first 6 time steps). Static features remain constant
across time.}
\label{tab:datasample}
\begin{tabular}{ccccccc}
\toprule
$t$ & $d_t^{(1)}$ & $d_t^{(2)}$ & $s_1$ & $s_2$ & $s_3$ & $s_4$ \\
\midrule
1 & -5.36 & 0.36 & 2 & 16.43 & 3 & 3 \\
2 & -4.52 & 6.41 & 2 & 16.43 & 3 & 3 \\
3 & -2.87 & 3.73 & 2 & 16.43 & 3 & 3 \\
4 & -1.38 & 2.48 & 2 & 16.43 & 3 & 3 \\
5 & -2.02 & 6.28 & 2 & 16.43 & 3 & 3 \\
6 & -1.76 & 2.45 & 2 & 16.43 & 3 & 3 \\
\bottomrule
\end{tabular}
\end{table}

The dynamic features exhibit non-linear and noisy temporal behaviour,
including trend and oscillatory components, while the static features
act as sequence-level conditioning variables that modulate the target
generation process. As shown in the dataset snippet, static features are fixed across time-steps within a sequence, the dynamic features, however, vary over time sequence. This structure reflects scenarios commonly encountered
in financial and economic time-series data, where asset-specific or
regime-specific factors remain constant over short horizons while market
signals evolve temporally.

\subsection{Compared Models}\label{iii.ii.-compared-models}
The proposed method is evaluated against representative baseline models
that capture different design paradigms commonly used in time-series
forecasting. In particular, we compare against a convolutional
sequence model and a patch-based Transformer model, which serve as
strong and widely adopted baselines for modelling local and global
temporal dependencies.

\paragraph{Temporal Convolutional Network (TCN)}
Temporal Convolutional Networks (TCNs) \cite{bai2018} model temporal
dependencies using stacks of one-dimensional dilated convolutions,
allowing large receptive fields without recurrent connections. Their
fully convolutional structure enables stable training and parallel
computation, making them a widely used baseline for time-series
forecasting. In our experiments, the TCN operates directly on the raw
multivariate input sequence and produces patch-aligned forecasts under a
comparable training protocol.

\paragraph{PatchTST}
PatchTST \cite{nie2023} is a Transformer-based forecasting model
that applies self-attention to fixed-length temporal patches rather than
individual time steps. By treating patches as tokens, PatchTST reduces
sequence length and focuses attention on patch-level temporal structure.
It serves as a strong baseline for evaluating patch-based Transformer
approaches in time-series forecasting.

All baseline models are trained and evaluated under the same
experimental conditions as the proposed method, using identical data
splits, input features, and evaluation metrics, to ensure a fair and
consistent comparison.
\subsection{Implementation
details}
\label{sec:hyperparams}

The proposed model and the compared models are implemented in Python using PyTorch; the architecture and experimentation code are publicly available \cite{updatecoderelease}. All models are trained using the AdamW optimizer. The training-validation split is 80:20 for the encoder and forecasting architectures. The CNN encoder is pretrained for 2,000 epochs for patch-level representation reconstruction, whereas the Transformer forecasting stage, TCN, and PatchTST are trained for 300 epochs. These epoch budgets correspond to different learning objectives and should not be interpreted as equal compute budgets.

\subsection{Hyperparameter Settings}

Tables \ref{tab:hyperparams} and \ref{tab:modelshyperparams} summarize the global and model-specific hyperparameter settings used in the experiments.

\begin{table}[!ht]
\centering
\caption{Key hyperparameter settings used in all experiments.}
\label{tab:hyperparams}
\begin{tabular}{lc}
\hline
\textbf{Parameter} & \textbf{Value} \\
\hline
Number of input features & 6 \\
Sequence length ($T$) & 160 \\
Patch size ($P$) & 8 \\
Number of patches ($\lfloor T/P \rfloor$) & 20 \\
Forecasting granularity & Patch-level \\
\hline
\end{tabular}
\end{table}

\begin{table*}[!ht]
\centering
\caption{Hyperparameter settings for the proposed model and baseline methods.}
\label{tab:modelshyperparams}
\begin{tabular}{lcccccc}
\toprule
\textbf{Model} & \textbf{Architecture} & \textbf{Layers} & \textbf{Attention Heads} & \textbf{Learning Rate} & \textbf{Epochs} & \textbf{Batch Size} \\
\midrule
\multirow{2}{*}{Proposed}
        & CNN Patch Encoder & -- & -- & $1\times10^{-3}$ & 2000 & 32 \\
        & Transformer Encoder & 4 & 16 & $1\times10^{-3}$ & 300 & 32 \\
\addlinespace
TCN & Temporal Conv. Network & 4 & -- & $1\times10^{-3}$ & 300 & 32 \\
\addlinespace
PatchTST & Patch-based Transformer & 4 & 16 & $1\times10^{-3}$ & 300 & 32 \\
\bottomrule
\end{tabular}
\end{table*}

\section{Results}\label{iv.-results}

The forecasting performance of the evaluated models is assessed using
Mean Squared Error (MSE) and Mean Absolute Error (MAE), defined as
\[
MSE=\frac1{NT} \sum^N_{i=1} \sum^T_{t=1} (y_{i,t}-\hat y_{i,t})^2,
\]
\[
MAE=\frac1{NT} \sum^N_{i=1} \sum^T_{t=1} \lvert y_{i,t}-\hat y_{i,t} \rvert,
\]
where \(N\) denotes the number of samples, \(T\) the sequence length,
\(y_{i,t}\) the ground-truth target value at time step \(t\) for sample
\(i\), and \(\hat y_{i,t}\) the corresponding predicted value. Both
metrics are computed at the patch level and averaged over all predicted
patches. 

\subsection{Synthetic Dataset}

The synthetic evaluation uses $N=10{,}000$ samples, each containing $T=160$ time steps and six features. With patch length $P=8$, each sequence is represented by $K=20$ patches. Results are reported at horizons $h\in\{1,3,5\}$, corresponding to 8, 24, and 40 time steps. All models are evaluated on held-out data generated independently from the training data. MSE, MAE, and RMSE are reported at patch level.

Table~\ref{tab:synthetic-results-updated} summarizes the results. At $h=1$, the proposed CNN--attention--Transformer framework achieves an MSE of 0.076 and MAE of 0.205. This is lower than the TCN's MSE of 0.238 and MAE of 0.355, and also lower than PatchTST's MSE of 0.095 and MAE of 0.218. At $h=3$, the proposed model records an MSE of 0.058860 and MAE of 0.190912, compared with 0.105902 and 0.218461 for TCN and 0.101054 and 0.242811 for PatchTST. At $h=5$, the proposed model records an MSE of 0.063722 and MAE of 0.197478, again below the reported baseline values.

The differences across horizons provide a more informative view than a single aggregate score. At the shortest horizon, all models have access to temporally nearby context, but the proposed encoder's local feature extraction and attention-weighted aggregation appear to provide useful patch representations. At longer horizons, the target is more distant from the input context, increasing the need to represent both local patterns and relationships across patches. The proposed model's lower errors in these experiments are consistent with the intended division of labour between the CNN encoder and Transformer forecasting stage. This interpretation is limited to the present synthetic data generator and selected baselines; it does not imply that the proposed method will outperform PatchTST on every dataset.

\begin{table*}[!ht]
\centering
\caption{Forecasting performance on the synthetic dataset. Lower values indicate better performance.}
\label{tab:synthetic-results-updated}
\footnotesize
\begin{tabular}{clrrr}
\toprule
\textbf{Horizon} & \textbf{Model} & \textbf{MSE} & \textbf{MAE} & \textbf{RMSE}\\
\midrule
\multirow{3}{*}{$h=1$ (8 steps)} & Proposed CNN--Attention--Transformer & 0.076000 & 0.205000 & 0.275681\\
& TCN & 0.238000 & 0.355000 & 0.487852\\
& PatchTST & 0.095000 & 0.218000 & 0.308221\\
\midrule
\multirow{3}{*}{$h=3$ (24 steps)} & Proposed CNN--Attention--Transformer & 0.058860 & 0.190912 & 0.242611\\
& TCN & 0.105902 & 0.218461 & 0.325426\\
& PatchTST & 0.101054 & 0.242811 & 0.317890\\
\midrule
\multirow{3}{*}{$h=5$ (40 steps)} & Proposed CNN--Attention--Transformer & 0.063722 & 0.197478 & 0.252433\\
& TCN & 0.158186 & 0.291809 & 0.397726\\
& PatchTST & 0.089389 & 0.224623 & 0.298980\\
\bottomrule
\end{tabular}
\end{table*}

\subsection{Real-World Evaluation: Electricity Load Forecasting}

To examine performance beyond a controlled synthetic setting, the proposed architecture is evaluated on the Electricity Load Diagrams 2011--2014 dataset from the UCI Machine Learning Repository \cite{electricity}. The dataset contains electricity consumption measurements collected at 15-minute intervals from individual customers over multiple years. Such data exhibit periodic structure, customer-specific consumption levels, short-term fluctuations, and irregular deviations from typical daily patterns. These properties make electricity load a useful test case for temporal representation learning.

The data are normalized on a per-meter basis and segmented into sliding windows of length $T=160$. The target is defined as the patch-level average load for the selected meter, with patch length $P=8$. This yields 20 complete patches per input window. The CNN patch encoder is reused without architectural modification, while the forecasting Transformer models the resulting sequence of patch representations. The use of the same patch granularity preserves consistency with the synthetic experiments, although the datasets differ in their feature distributions and temporal structure.

Table~\ref{tab:electricity-results-updated} presents the reported test-set performance. The proposed model achieves an MSE of 0.090665, an MAE of 0.142513, and an RMSE of 0.301105. The corresponding TCN values are 0.145150, 0.195749, and 0.380985, while PatchTST obtains 0.128451, 0.187312, and 0.358400. Thus, the proposed model has the lowest reported error across all three metrics in this experiment. Relative to PatchTST, its MSE is approximately 29.4\% lower; relative to TCN, it is approximately 37.5\% lower.

\begin{table}[!ht]
\centering
\caption{Test-set performance on Electricity Load Diagrams 2011--2014.}
\label{tab:electricity-results-updated}
\footnotesize
\begin{tabular}{lrrr}
\toprule
\textbf{Model} & \textbf{MSE} & \textbf{MAE} & \textbf{RMSE}\\
\midrule
Proposed model & 0.090665 & 0.142513 & 0.301105\\
TCN & 0.145150 & 0.195749 & 0.380985\\
PatchTST & 0.128451 & 0.187312 & 0.358400\\
\bottomrule
\end{tabular}
\end{table}

\subsection{Ablation Study}

An ablation study was conducted on the synthetic dataset to examine the role of three representation components: dense CNN connections, attention-weighted pooling, and token-level self-attention. The full configuration is compared with variants that remove dense connections, replace attention-weighted pooling with mean pooling, or remove token self-attention. The purpose is to determine whether the full pipeline's performance can be attributed to a single component or whether the components provide complementary benefits.

\begin{table}[!ht]
\centering
\caption{Ablation results on the synthetic dataset. Lower values indicate better performance.}
\label{tab:ablation-updated}
\footnotesize
\begin{tabular}{lrrr}
\toprule
\textbf{Configuration} & \textbf{MSE} & \textbf{MAE} & \textbf{RMSE}\\
\midrule
Full model & 0.097670 & 0.222652 & 0.312522\\
No dense connections & 1.304764 & 0.781693 & 1.142263\\
Mean pooling (no attention pool) & 2.029864 & 0.810744 & 1.424733\\
No token attention & 0.546755 & 0.445993 & 0.739429\\
\bottomrule
\end{tabular}
\end{table}

The full model obtains the lowest error among the evaluated ablations. Removing dense connections increases MSE from 0.097670 to 1.304764, suggesting that preserving intermediate convolutional features is useful in this configuration. Replacing attention-weighted pooling with mean pooling produces the largest degradation, with MSE increasing to 2.029864. This indicates that learned temporal aggregation is important when local observations contribute unequally to the patch representation. Removing token-level self-attention increases MSE to 0.546755, suggesting that token refinement contributes additional information beyond the CNN and pooling stages.

\section{Conclusion}\label{conclusion}

This study presented a two-stage framework for multivariate time-series forecasting that separates local patch representation learning from global inter-patch modelling. A shared CNN encoder extracts local temporal patterns from non-overlapping patches, dense connections preserve intermediate feature information, attention-weighted pooling aggregates within-patch activations, and token-level self-attention refines the patch sequence before a Transformer encoder generates forecasts. Patch targets are explicitly defined using mean reduction over complete patches, providing a consistent forecasting granularity.

On the synthetic dataset, the proposed model achieved the lowest reported MSE and MAE among the evaluated TCN and PatchTST baselines at horizons of one, three, and five patches. On the Electricity Load Diagrams evaluation, it also achieved lower MSE, MAE, and RMSE than the reported baselines. The ablation study showed substantial degradation when dense connections, attention-weighted pooling, or token-level self-attention were removed, with mean pooling producing the largest error increase. These results report the performance of the proposed framework on the synthetic dataset and the Electricity Load Diagrams evaluation.

\section*{Acknowledgment}
The authors acknowledge the support provided by VIT-AP University during the course of this research.

\end{document}